\documentclass[11pt]{article}

\usepackage[margin=1in]{geometry}
\usepackage{amsmath,amssymb,amsthm}
\usepackage{mathtools}
\usepackage{enumitem}
\usepackage{xcolor}
\usepackage{graphicx}
\graphicspath{{formalize/retrospective/}}
\usepackage[section]{placeins}   
\usepackage{needspace}             
\usepackage{fontspec}              
\usepackage{stmaryrd}              
\usepackage{newunicodechar}        
\newunicodechar{⨆}{\ensuremath{\bigsqcup}}
\newunicodechar{⨅}{\ensuremath{\bigsqcap}}
\usepackage{fancyvrb}
\DefineVerbatimEnvironment{leancode}{Verbatim}%
  {frame=single,framesep=4pt,fontsize=\small,xleftmargin=3pt,xrightmargin=3pt,%
   codes={\catcode`⨆=\active \catcode`⨅=\active}}  
\usepackage{hyperref}
\usepackage{xurl}  
\hypersetup{colorlinks=true,linkcolor=blue!50!black,citecolor=blue!50!black,urlcolor=blue!50!black}

\theoremstyle{plain}
\newtheorem{theorem}{Theorem}[section]

\newtheorem{corollary}[theorem]{Corollary}
\theoremstyle{definition}
\newtheorem{definition}[theorem]{Definition}

\newtheorem{assumption}{Assumption}
\theoremstyle{remark}

\newcommand{\R}{\mathbb{R}}

\newcommand{\Wone}{W_1}
\newcommand{\Wbar}{\overline{W}}
\newcommand{\lean}[1]{\texttt{#1}}
\newcommand{\leanbaseurl}{https://hydrodynamical.github.io/Vlasov\_Meanfield\_Formalization/docs/find/\#doc}
\newcommand{\leanlink}[1]{\ifx\leanbaseurl\empty\lean{#1}\else\href{\leanbaseurl/Vlasov.#1}{\lean{#1}}\fi}
\newcommand{\printaxioms}{\lean{\#print axioms}}
\newcommand{\cleanfootprint}{\lean{[propext, Classical.choice, Quot.sound]}}

\title{\textbf{A Formalization of the Mean-Field Derivation\\ of the Vlasov Equation}\\[2pt]
\large Mathematician in the loop: AI-assisted Lean formalization as a strategy game}

\author{Joseph K. Miller\thanks{%
  Stanford University, \href{mailto:jkm314@stanford.edu}{jkm314@stanford.edu};
  Massachusetts Institute of Technology, \href{mailto:jkm314@mit.edu}{jkm314@mit.edu}.
  The development is public. Source:
  \url{https://github.com/Hydrodynamical/Vlasov_Meanfield_Formalization}. Blueprint site
  and API documentation:
  \url{https://hydrodynamical.github.io/Vlasov_Meanfield_Formalization/}.}}

\date{\today}

\begin{document}
\maketitle

\begin{abstract}
We formalize a research result in the Lean~4 proof assistant by having a
mathematician direct an AI system, and frame the activity as a \emph{formalization game}. The
objective is to turn a \LaTeX{} document into Lean. The game is \emph{won} when the
development compiles, contains no \lean{sorry}, and a machine check shows the target
theorems rest on Lean's foundational axioms alone. Reuse is a second check, by a
definition we introduce: whether the development yields a \emph{self-contained layer}
of general mathematics the wider library could absorb.

The case study is a complete, axiom-clean formalization of well-posedness for the
nonlinear Vlasov equation via Dobrushin's mean-field route --- existence, uniqueness, the
stability estimate and mean-field limit, and a short-window superposition principle
(weak solutions are Lagrangian). The human's
role here was to direct, not to write proofs: to scope the definitions, steer the
decompositions, and triage the library's gaps; the AI agent executed. The formalization certifies the
proof of each statement as written; whether the written statement is the intended
theorem stays the mathematician's judgment. The optimal-transport
machinery that fell out of the build (in particular, properties of the Wasserstein-$1$
metric and the Kantorovich--Rubinstein duality theorem) separates into a self-contained layer
(Definition~\ref{def:selfcontained}) that compiles against Mathlib alone: about a
sixth of the development ($49$ of $299$ declarations), behind a $22$-declaration
interface with no reverse dependency. The headline theorems ran in about a week, the
full development in about a month. We report
the quantitative claims as observations of one game, not as general laws. The game's
rules name no particular system, so the methodological framing is meant to outlast the tools of any
one run.
\end{abstract}

\tableofcontents

\section{Introduction}

A proof on paper is an argument that asks a reader to be convinced. A proof in a
formal proof assistant is a different kind of object: one the computer checks. The
definitions, the theorem, and the proof are written in a formal language, and the
machine either accepts the proof or it does not. There is no room for ``clearly,''
and no referee to persuade. Lean is one such language. Over the past decade a
community has assembled in it a large and growing library of mathematics, called
Mathlib, on which new formal proofs can be built.

For most of its history, formalization has trailed far behind the mathematics it
certifies. A result a mathematician reads in an afternoon can take a skilled
formalizer weeks to put into Lean, and the obstacle is rarely the logic. It is the
volume of small intermediate steps, the labor of matching an informal argument to
the library's conventions, and, in analysis especially, the parts of the library
that do not yet exist. This is what has kept formal proof a specialist's craft
rather than an everyday tool.

Two things have recently changed: AI systems can now write Lean proofs, and they
can write them quickly. This raises the possibility that formalization can
keep pace with ordinary mathematical work. That speed sharpens two questions about what the machine produces: whether one can
trust proofs one did not write, and whether anyone can use what is left behind. This
paper turns both into machine checks. Trust is answered by a \emph{win condition} --- the
development compiles, contains no \lean{sorry}, and depends on Lean's foundational
axioms alone --- which certifies a proof relative to the statement as written, leaving
statement-faithfulness with the mathematician. Reuse is answered by a
\emph{self-contained layer} --- a dependency-isolated body of general mathematics,
behind a small interface, that builds against the ambient library alone --- which
certifies that what remains is absorbable rather than disposable. Both are stated
precisely in Section~\ref{sec:game} (Definitions~\ref{def:win}
and~\ref{def:selfcontained}).

We put both checks to the test by doing the work and reporting what it required. We
formalized a genuine result from kinetic theory --- Dobrushin's derivation of the nonlinear
Vlasov equation, a classical passage from the motion of many interacting particles to
the equation governing their average --- in Lean, over about a month. We did not write the Lean
ourselves: a human mathematician directed an AI system, which wrote the proofs, while
the mathematician decided what to prove, how to cut each hard theorem into reachable
pieces, and when to abandon a line that was not working.

The win condition was met here, and the development yields a self-contained layer. 
The first is discharged by the machine: \printaxioms{} reports
the target theorems depend on Lean's foundations alone. What it cannot supply is the
other half --- that each statement as written is the theorem intended --- nor the
judgment that carried the proofs; the case study marks both, stage by stage. Getting
the solution concept right (Section~\ref{ssec:newton-to-weak}): the weak-solution
test class had silently collapsed, leaving a key statement vacuously true behind a
green build, caught by a human read. Building the distance
(Section~\ref{ssec:wasserstein}): the duality bridging its two faces was absent from
the library, won by human-steered search and agent throughput. The stability
estimate (Section~\ref{ssec:dynamical-core}): read in the dual face it appears to
need an attainment theorem the library lacks; a human redirection, back to
Dobrushin's own coupling formulation, removed the need. The superposition principle
(Section~\ref{ssec:superposition}): one degree of regularity deliberately spent to
make uniqueness go through --- a human typeclass decision, the construction the
agent's. The layer is delivered in turn: the general mathematics the argument
demanded separates cleanly from the problem-specific development
(Section~\ref{ssec:reuse}), fit for the ambient library. The human side of each
unlock is not incidental --- that judgment accumulates, in a standing guidance file
built as play proceeds (Section~\ref{sec:strategies}).

\subsection{Related work}
This work borders three recent lines: the end-to-end formalization of known-hard
results, the autonomous proof-search agents, and first-hand accounts
of AI-assisted mathematics.

The first is the now-established tradition of formalizing a substantial,
known-hard mathematical result \emph{end to end} in Lean: the Liquid Tensor Experiment
\cite{LTE}, the polynomial Freiman--Ruzsa theorem \cite{PFR}, Carleson's theorem
on the convergence of Fourier series \cite{Carleson}, and the sphere-eversion
project \cite{SphereEversion} are the canonical examples. These share an
infrastructure --- Massot's \lean{leanblueprint} \cite{leanblueprint}, a
human-readable proof skeleton linked to the formal development, which the present
project also uses --- and a common mode: a human mathematician, or team, directs
the formalization of a result whose mathematics is already understood. The present
work is squarely in that lineage; what sets it apart is making the \emph{strategy}
of the human direction explicit --- naming, phase by phase, what the mathematician
is doing.

A parallel and recent line is \emph{autonomous}. AlphaProof \cite{AlphaProof}
couples reinforcement learning with Lean and reached silver-medal performance at
the International Mathematical Olympiad; its agentic successor
\cite{AlphaProofNexus} and Harmonic's Aristotle \cite{Aristotle} drive
evolutionary language-model/Lean loops and have autonomously resolved open
Erd\H{o}s problems \cite{Erdos728}; LeanMarathon \cite{LeanMarathon} is a
multi-agent, blueprint-centric harness that autoformalizes the theorems of recent
Erd\H{o}s-problem papers; and Alexeev, Lichtman, Tao, and coauthors
\cite{PrimitiveSets} resolved open Erd\H{o}s conjectures on primitive sets with
a key method suggested by output of GPT-5.4~Pro and the results formalized in
Lean, in part by Math~Inc.'s autonomous Gauss. The present work runs at the
other end of the autonomy axis: the mathematician directs throughout, and
\emph{documenting} that direction --- where to cut a hard theorem, which routes
the ambient library can support, when to abandon one --- is itself a facet of
the paper. The autonomous systems do exhibit exactly the failure mode the win condition
(Definition~\ref{def:win}) is built to catch --- a renamed \lean{sorry}, a hallucinated
lemma cited as settled --- documented by the systems' own authors \cite{AlphaProofNexus}.

Closest to this paper are \emph{first-hand} accounts of AI-assisted
research mathematics. Ilin \cite{IlinVML} semi-autonomously formalized a
Vlasov--Maxwell--Landau \emph{equilibrium} in Lean~4 under a single supervising
mathematician (using, among other tools, Claude Code and Aristotle) --- the same
domain and broadly the same mode as here, but a static characterization rather
than the dynamic well-posedness, stability, the limit, and
superposition principle \cite{AGS} we formalize, and without the methodological
apparatus we foreground. In mathematical physics more broadly, Douglas, Hoback,
Mei, and Nissim \cite{QFTFormalization} formalized the free bosonic quantum field
theory in four dimensions against the Glimm--Jaffe axioms in Lean~4, as a proof of
concept for AI-assisted formalization of extended mathematical-physics arguments;
their original release assumed three classical results (Minlos' theorem among
them) as disclosed placeholders --- the deferred gap of Section~\ref{sec:phases},
since discharged. Contemporaneously, Armstrong and Kempe \cite{DGNM}
formalized the core interior De Giorgi--Nash--Moser regularity theory for
uniformly elliptic divergence-form equations in Lean~4 --- to their knowledge the
first machine-checked formalization of a major theorem in modern PDE regularity ---
likewise carried out with extensive use of large language models under close
human supervision. Their account, like ours, locates the decisive human
contributions in detailed proof blueprints, careful interface design, and
validation against the checked artifact.
In harmonic analysis, de~Dios~Pont, Taylor, and coauthors have posted a series
of papers pairing new theorems with Lean~4 formalizations: stable phase
retrieval in two settings \cite{STFTPhaseRetrieval,PhaseRetrievalSpans}, a
negative answer to the existence problem for regular Gabor frames
\cite{GaborFrames}, and an answer to a question of Strichartz on Fourier frames
for Cantor measures \cite{CantorFourierFrames}. There the formalization
certifies mathematics that is new --- two of the papers
\cite{STFTPhaseRetrieval,PhaseRetrievalSpans} report proofs developed in
interplay with LLMs --- and the last of the series was posted the same day as
the present paper.
Zimmer et al.\ \cite{AgenticResearcher} encode agentic-research norms as prompt
``commandments,'' a close parallel to our standing-instruction file; and Avigad
\cite{Avigad} argues that mathematicians should actively deploy
and shape these tools rather than merely react to them --- a call to which this
paper is a concrete response. What distinguishes the present work against this
backdrop is the target it runs on: a \emph{dynamical} result for a nonlinear
PDE --- well-posedness, stability, the mean-field limit, superposition ---
carried end to end as one development.

\subsection{Choosing the target}
Derivations of this kind come in a range of difficulty, set by how rough the force
between particles is. Where the force is smooth, the passage to the limiting
equation is classical, going back to Dobrushin \cite{Dobrushin} and the
Braun--Hepp--Neunzert line \cite{BraunHepp,Neunzert} (with the later surveys
\cite{Spohn,Golse}). Where the force is singular, as in gravitation or
electrostatics, the same passage from the second-order Newtonian dynamics is a
celebrated open problem, settled so far only in softened cases
\cite{hauray-jabin,lazarovici-pickl,feistl-held-pickl,feistl-held-pickl-2d};
its first-order analogue is settled for Coulomb and Riesz interactions
\cite{serfaty-coulomb,nguyen-rosenzweig-serfaty,nguyen-serfaty}. We formalize the smooth
end, and chose it deliberately: it is where a reusable library can be built. Closing
the singular frontier is a problem of mathematics, not of proof engineering --- the
missing ingredient is new analysis, and the most a mature library could offer there
is a place to check such arguments as they are found. The library amplifies the
mathematician; it does not stand in for the mathematics.

We took it less for any one theorem than as a faithful representative of its kind:
its stages --- the empirical distribution of the particles, a notion of distance
between such distributions, the flow that moves them, and the equation they satisfy
in the limit --- recur across the subject. Writing it in Lean meant first building
general mathematics the library does not yet contain, about distances between
probability distributions and the flows that move them. To our knowledge no prior
formalization had assembled it; this is the self-contained layer of
Definition~\ref{def:selfcontained}, and the lasting artifact of the exercise.

We describe the activity as a \emph{formalization game} because the framing is precise enough to be load-bearing.
There is a definite objective (turn the \LaTeX{} into Lean), a machine-checkable
win condition (no \lean{sorry}, a clean axiom footprint), and a payoff beyond
winning --- whether the won development yields a \emph{self-contained layer}. And there are three
phases, each rewarding a different strategy: a player who opens well can still lose
the endgame. The framing is also built to outlast its tools: the rules name no
particular system and no division of labor, so what was attempted and what was won
stays meaningful as the machines underneath turn over
(Section~\ref{sec:assessment}).

\subsection{What we formalize}\label{sec:statements}
\begin{figure}[t]
\centering
\includegraphics[width=\textwidth]{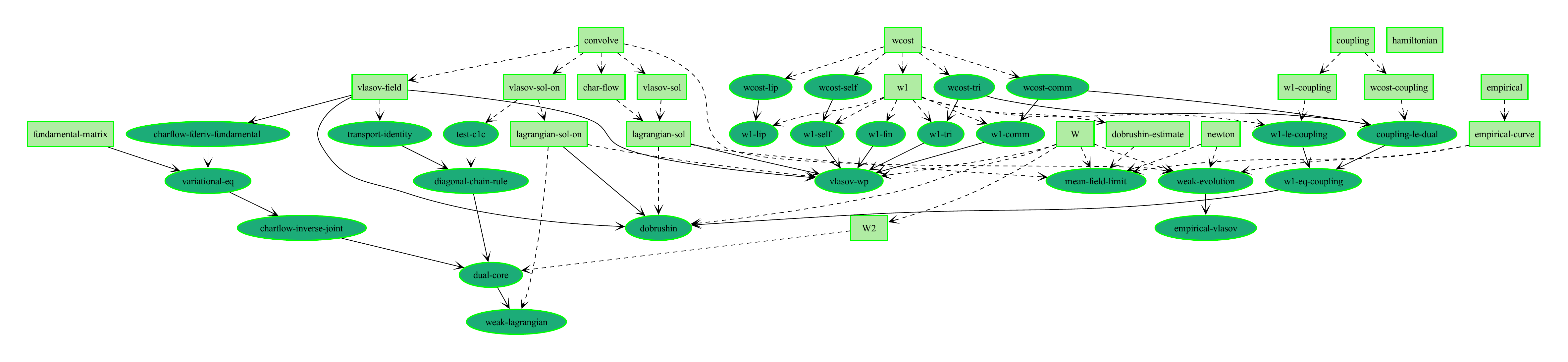}
\caption{A roadmap of the development: the project blueprint's dependency graph,
auto-extracted from the elaborated Lean proof terms --- boxes are definitions,
ellipses theorems, green fully formalized. The interactive version, with statements
and proof sketches on every node, is at
\href{https://hydrodynamical.github.io/Vlasov_Meanfield_Formalization/}{\nolinkurl{hydrodynamical.github.io/Vlasov_Meanfield_Formalization}}.}
\label{fig:depgraph}
\end{figure}

We now make that passage precise. The
particles carry positions $x_i$ and velocities $v_i$ and evolve under the
pairwise force $-\nabla W$ by Newton's equations
\begin{equation}\label{eqn:Newton-Intro}
  \dot{x}_i \;=\; v_i, \qquad
  \dot{v}_i \;=\; -\frac{1}{N} \sum_{j \neq i} \nabla W(x_i - x_j),
  \qquad i = 1, \dots, N .
\end{equation}
As $N \to \infty$ the empirical distribution of the particles concentrates on a probability density $f(t,x,v)$ on $\R^d \times \R^d$ governed by the nonlinear Vlasov equation
\begin{equation}\label{eqn:Vlasov-Intro}
  \partial_t f \;+\; v \cdot \nabla_x f
    \;-\; (\nabla W * \rho_t)(x) \cdot \nabla_v f \;=\; 0,
  \qquad \rho_t(x) = \int_{\R^d} f(t,x,v)\, dv ,
\end{equation}
with $\rho_t$ the spatial marginal. Figure~\ref{fig:depgraph} is a roadmap of what follows --- the statements below and the layered development that proves them. Here we give the mathematics; the verbatim Lean signatures are collected in Appendix~\ref{app:formal}.

The standing assumption, in force throughout, concerns the regularity of the interaction potential.

\begin{assumption}[\leanlink{AssW}]\label{ass:1}
  Let $W: \R^d \rightarrow \R$ be an even function that is $C^{1,1}$, with a globally Lipschitz gradient of constant $L := \mathrm{Lip}(\nabla W)$. 
\end{assumption}

Write $\mathcal{P}_1(\R^d\times\R^d)$ for the Borel probability measures $\mu$ on phase space with finite first moment, $\int_{\R^d\times\R^d}(|x|+|v|)\,d\mu(x,v)<\infty$; in Lean this is the predicate \lean{HasFiniteFirstMoment}. A solution of \eqref{eqn:Vlasov-Intro} comes in two forms, and we begin with the \emph{weak} (Eulerian) one. Each notion below is a property of the whole curve $t\mapsto f_t$; its \emph{initial datum} is the value $f(0)$, pinned to a prescribed $f_0$ only when we solve the Cauchy problem (Theorem~\ref{thm:wp}).

\begin{definition}[\leanlink{IsVlasovSolution}]\label{def:weak-Vlasov}
  A \emph{weak} solution of \eqref{eqn:Vlasov-Intro} on $[0,T]$, $T>0$, is any $f : [0,T] \rightarrow \mathcal{P}_1(\R^d \times \R^d)$ which is narrowly continuous in time (continuous against bounded continuous test functions) and satisfies, for all $\varphi\in \mathcal{C}^\infty_c(\R^{d}\times \R^d)$:
  \begin{itemize}
    \item The map $t\mapsto \langle f_t , \varphi\rangle$ is differentiable, and 
    \item The derivative satisfies 
    \begin{equation*}
      \frac{d}{dt}\langle f_t,\varphi\rangle = \langle f_t,\, v\cdot\nabla_x\varphi -
(\nabla W * \rho_t)\cdot\nabla_v\varphi\rangle .
    \end{equation*}
  \end{itemize}
\end{definition}

An alternative notion of solution of \eqref{eqn:Vlasov-Intro} is a \emph{Lagrangian} solution: a weak solution transported by its own characteristic flow.

\begin{definition}[\leanlink{IsLagrangianVlasovSolution}]\label{def:Lagrangian}
  A \emph{Lagrangian} solution is a
  weak solution $f:[0,T]\to\mathcal{P}_1(\R^d\times\R^d)$ for which there exists a flow
  $(X,V):[0,T]\times(\R^d\times\R^d)\to\R^d\times\R^d$ solving the characteristic system
  \[
    \dot X(t,z)=V(t,z),\qquad \dot V(t,z)=-(\nabla W*\rho_t)\big(X(t,z)\big),\qquad (X,V)(0,z)=z,
  \]
  such that $f_t=\big(X(t,\cdot),V(t,\cdot)\big)_\#\,f(0)$ for every $t\in[0,T]$. Here the
  pushforward $\Phi_\#\mu$ of a measure $\mu$ by a Borel map $\Phi$ is
  $(\Phi_\#\mu)(A)=\mu\big(\Phi^{-1}(A)\big)$, equivalently $\int \varphi\,d(\Phi_\#\mu)=\int
  \varphi\circ\Phi\,d\mu$ for every bounded continuous $\varphi$.
\end{definition}

\noindent The Lean predicates are leaner than the prose: \leanlink{IsVlasovSolution} states
exactly the differentiability and the derivative identity, for a curve of measures over all
of $\R$, and \leanlink{IsLagrangianVlasovSolution} adds the flow witness
(Appendix~\ref{app:formal}); membership in $\mathcal{P}_1$, the window, and continuity in
time are imposed where they are consumed, as explicit hypotheses of the theorems.
Theorem~\ref{thm:superposition} quantifies over the windowed variant
\lean{IsVlasovSolutionOn}, with continuity in time tested against compactly supported
continuous functions --- equivalent to narrow continuity for probability-valued curves.

Every Lagrangian solution is weak, by definition. The converse --- that every weak solution
is Lagrangian --- is the substantive direction: the classical superposition principle \cite{AGS}. We
prove a short-window version under the strengthened assumption $\mathrm{AssW2}$ introduced below
(Theorem~\ref{thm:superposition}; Section~\ref{ssec:superposition}).

\begin{theorem}[Global well-posedness, \leanlink{vlasovWellPosedness}]\label{thm:wp}
  Let $W$ satisfy Assumption~\ref{ass:1} and $f_0 \in \mathcal{P}_1(\R^d \times \R^d)$. The forward
  Cauchy problem is well-posed: there exists a \emph{single} curve
  $f:[0,\infty)\rightarrow \mathcal{P}_1(\R^d \times \R^d)$ with datum $f(0)=f_0$ and finite first
  moment at every $t\ge 0$, which is a Lagrangian solution of \eqref{eqn:Vlasov-Intro} on every
  window $[0,T]$, in the sense of Definitions~\ref{def:weak-Vlasov} and~\ref{def:Lagrangian}; and on
  each window it is the unique such Lagrangian solution.
\end{theorem}
\noindent The single global curve is \leanlink{vlasovWellPosedness}; per-window uniqueness in the
class of Definition~\ref{def:Lagrangian} is \leanlink{vlasovWellPosedness\_uniqueness}, a separate
statement. It follows for $L>0$ from the stability estimate below; the degenerate $L=0$ case
(constant force) is explicit, so uniqueness holds at every $L$.

Solutions of \eqref{eqn:Vlasov-Intro} depend stably on their data; the relevant distance is the Wasserstein-1
(Kantorovich--Rubinstein) metric on $\mathcal{P}_1$.
\begin{definition}[\leanlink{wasserstein1}]\label{def:w1}
  For $\mu,\nu \in \mathcal{P}_1(\R^d\times\R^d)$,
  \[
    \Wone(\mu,\nu) \;:=\; \sup_{\mathrm{Lip}(f)\le 1}\Big(\int f\,d\mu - \int f\,d\nu\Big)
    \;=\; \inf_{\pi\in\Pi(\mu,\nu)} \int \mathrm{dist}(z_1,z_2)\,d\pi ,
  \]
  the supremum over $1$-Lipschitz $f:\R^d\times\R^d\to\R$ and the infimum over the set
  $\Pi(\mu,\nu)$ of \emph{couplings} --- probability measures on the product with marginals
  $\mu$ and $\nu$. The supremum is the \emph{dual} face, the infimum the \emph{primal};
  they agree by Kantorovich--Rubinstein duality \cite{Villani,AGS}.
\end{definition}
\noindent Both faces and the duality between them are proved in the development: the easy
direction (dual $\le$ primal) is \leanlink{wasserstein1\_le\_wasserstein1\_coupling}; the
equality, for probability measures with finite first moment on a Polish space, is
\leanlink{wasserstein1\_eq\_coupling}, whose hard direction is built in
Section~\ref{ssec:wasserstein}. Neither optimum is attained anywhere --- every bound is
$\varepsilon$-optimal.
\begin{theorem}[Dobrushin stability \cite{Dobrushin}, \leanlink{dobrushin}]\label{thm:dob}
  Let $W$ satisfy Assumption~\ref{ass:1}. Any two Lagrangian solutions $f,g$ with
  finite first moment at every time are stable in $\Wone$ at an exponential rate,
  \[ \Wone(f_t,g_t)\;\le\; e^{Ct}\,\Wone(f_0,g_0),\qquad t\ge 0,\quad C=C(L). \]
\end{theorem}

\noindent Applied with $g$ the empirical measure of the Newton dynamics
\eqref{eqn:Newton-Intro}, the estimate gives the mean-field limit as a corollary.
\begin{corollary}[Mean-field limit, \leanlink{meanFieldLimit}]\label{cor:mfl}
  Let $f$ be the Vlasov solution with datum $f_0$, let $\mu^N_t$ be the empirical measure of
  $N$ particles evolving by \eqref{eqn:Newton-Intro}, and assume the estimate of
  Theorem~\ref{thm:dob} holds for every pair $(\mu^N,f)$ with one constant $C$. If the initial
  empirical measures converge, $\Wone(\mu^N_0,f_0)\to0$ as $N\to\infty$, then for every $T>0$
  \[ \sup_{t\in[0,T]}\Wone(\mu^N_t,f_t)\;\xrightarrow[N\to\infty]{}\;0. \]
\end{corollary}
\noindent Theorem~\ref{thm:dob} supplies the assumed estimate --- each empirical curve is a
Lagrangian solution with finite moments, and the rate $C=2\max(1,L)$ of
Section~\ref{ssec:dynamical-core} is uniform in $N$ --- but the Lean takes it as an explicit
hypothesis rather than re-deriving it (Appendix~\ref{app:formal}).

Extending stability and uniqueness from Lagrangian solutions to \emph{all} weak solutions is the
superposition principle; the version we prove costs one degree of regularity beyond
Assumption~\ref{ass:1} and runs on a short window.
\begin{assumption}[\leanlink{AssW2}]\label{ass:2}
  In addition to Assumption~\ref{ass:1}, let $\nabla W \in C^1$ (equivalently $W \in C^2$).
\end{assumption}
\begin{theorem}[Superposition principle, \leanlink{weak\_isLagrangianVlasovSolutionOn}]\label{thm:superposition}
  Let $W$ satisfy Assumption~\ref{ass:2} and let $T>0$ satisfy $L\,T^{2}<1$. Every weak solution
  $f:[0,T]\to\mathcal{P}_1(\R^d\times\R^d)$ whose first moments are uniformly bounded on $[0,T]$
  and whose force field is regular --- $(\nabla W*\rho_t)(x)$ continuous in $t$ for each $x$, its
  spatial derivative jointly continuous on $[0,T]\times\R^d$ --- is Lagrangian on $[0,T]$.
\end{theorem}
\noindent The hypotheses, reproduced in full in Appendix~\ref{app:formal}, are not cosmetic: the
window is short, and the continuity hypotheses concern the solution's \emph{own} force field.
Chaining short windows to arbitrary $T$, as the well-posedness construction does for its flow,
is not formalized for this bridge; it is future work.

This introduction has stated the two checks --- the win condition and the
self-contained layer --- set down in mathematics
exactly what we formalized, and placed the work among its neighbors; the rest of
the paper is the method and its evidence. Section~\ref{sec:game} sets out the game
--- its win condition, its payoff (the self-contained layer), the three phases, and the
$\mathrm{L}/\mathrm{P}/\mathrm{M}/\mathrm{B}$ framework of standing guidance the
human accumulates. Section~\ref{sec:casestudy}
is the heart of the paper: Dobrushin's derivation, run as a formalization in its
mathematical order --- from the particle system to the limiting equation, through the
core estimates, to the gaps where the library ran out --- marking at each stage where
the mathematics was hard and who unlocked it. Section~\ref{sec:assessment} assesses the
method: the discipline that kept it sound at speed, the timeframe and division of labor,
what generalizes, and the limits of one run, closing with the next target and
with what should outlast the tools.

\section{The formalization game}\label{sec:game}

\subsection{Objective, win condition, self-contained layer}

The substance of the \emph{formalization game} is its win condition and the
self-contained layer a win can export. It is played on three objects: a source mathematics document (a
\LaTeX{} paper), a Lean~4 development $\mathcal{L}$ under construction, and a
standing instruction file $\mathcal{G}$ governing the agent's behavior. A
distinguished set of \emph{target theorems} $\Theta \subseteq \mathcal{L}$ encodes
the main results of the source, and the human and the agent alternate moves that
edit $\mathcal{L}$ and $\mathcal{G}$.

\begin{definition}[Win condition]\label{def:win}
The game is \emph{won} when:
\begin{enumerate}[label=(W\arabic*),leftmargin=3em]
\item $\mathcal{L}$ compiles, with no occurrence of \lean{sorry} in any
declaration reachable from $\Theta$; and
\item for every $\theta \in \Theta$, the command \printaxioms{}~$\theta$
returns exactly \cleanfootprint.
\end{enumerate}
\end{definition}

The second clause is the substantive one. A development can be free of
\lean{sorry} and still lean on a bespoke \lean{axiom} --- a \lean{sorry} renamed
to look principled. \printaxioms{} traces the \emph{complete} axiom set a theorem
depends on; a footprint of exactly \lean{propext}, \lean{Classical.choice}, and
\lean{Quot.sound} --- Lean's own foundations --- means proved modulo those alone,
and any fourth name is a loss a build-success check would miss. The win condition
is ``compiles \emph{and} the foundations are clean,'' not ``compiles.''

This certifies a theorem soundly \emph{relative to its statement as written}; it does
not certify that the written statement is the theorem intended. That second responsibility,
statement-faithfulness, is the human's, discharged in the early game through
definitional scoping and atom-level signature reading.

We define reusability as the existence of a \emph{self-contained layer}: a
subset of the development the ambient library could absorb with no trace of the
problem it came from, checked pass/fail by criteria the library defines, not us.

\begin{definition}[Self-contained layer]\label{def:selfcontained}
A won game exports one or more \emph{self-contained layers}. A subset
$\mathcal{L}_{\mathrm{gen}} \subseteq \mathcal{L}$ is \emph{self-contained} when
it passes four checks, each re-runnable by a referee and each defined by the
ambient library rather than by us:
\begin{enumerate}[label=(Q\arabic*),leftmargin=3em]
\item \textbf{Isolation.} In the elaboration environment's declaration-level
dependency graph, no edge runs from $\mathcal{L}_{\mathrm{gen}}$ into
$\mathcal{L} \setminus \mathcal{L}_{\mathrm{gen}}$: every cross-edge points
inward, through an interface of width $w$ (the number of
$\mathcal{L}_{\mathrm{gen}}$ declarations referenced from outside). The
reverse-edge count is $0$, read directly off the graph.
\item \textbf{Standalone compilation.} $\mathcal{L}_{\mathrm{gen}}$ builds as a
package depending only on the Mathlib library (v4.29.1), with zero
problem-specific files in its import graph. This is the compiled certificate of
(Q1): the build cannot succeed if a hidden reverse edge exists.
\item \textbf{Linter-clean.} $\mathcal{L}_{\mathrm{gen}}$ passes the ambient
library's own linter suite with zero warnings --- the automated gate its
maintainers apply before any merge (unused hypotheses, simp-normal-form, naming,
docstring coverage).
\item \textbf{Non-redundant.} No declaration in $\mathcal{L}_{\mathrm{gen}}$
restates a result already present in the ambient library, and every declaration
is reachable from a target theorem --- no duplication of existing mathematics, no
dead code.
\end{enumerate}
A layer passing (Q1)--(Q4) is \emph{self-contained}, separated from the
problem-specific development through an interface of width $w$. The \emph{size}
of $\mathcal{L}_{\mathrm{gen}}$, as a fraction of $\mathcal{L}$, is reported
descriptively, not as a criterion. We claim reusability in this precise, certified
sense, and distinguish it from demonstrated \emph{reuse}, which we do not assert.
\end{definition}

In the case
study, $\mathcal{L}_{\mathrm{gen}}$ is the optimal-transport machinery built for the
derivation --- the Wasserstein-1 metric in both faces, the
Kantorovich--Rubinstein bridge between them, a finite Kantorovich duality, and the
coupling-gluing triangle inequality (Table~\ref{tab:lgen}). It passes all four checks;
the numbers --- interface width, layer size, and the reverse-edge count --- are reported
with the outcome in Section~\ref{ssec:reuse}.

\begin{table}[t]
\centering
\begin{tabular}{ll}
\hline
General object (mathematics) & Lean declaration \\
\hline
Wasserstein-1, dual (Kantorovich--Rubinstein) form & \lean{wassersteinCost}, \lean{wasserstein1} \\
Wasserstein-1, primal (coupling) form & \lean{wasserstein1\_coupling} \\
Truncated variant $\Wbar$ (bounded ground cost) & \lean{wassersteinBar} \\
Dual $=$ primal (Kantorovich--Rubinstein bridge) & \lean{wasserstein1\_eq\_coupling} \\
Finite Kantorovich duality (from Farkas) & \lean{finiteRange\_transportation\_dual} \\
Coupling-cost triangle inequality (gluing) & \lean{wassersteinCost\_coupling\_triangle} \\
Non-expansion under Lipschitz pushforward & \lean{wasserstein1\_pushforward\_le\_iInf} \\
\hline
\end{tabular}
\caption{Representative contents of the self-contained layer
$\mathcal{L}_{\mathrm{gen}}$ --- general optimal-transport mathematics that
fell out of the build, stated once and reused. The full layer is $49$
declarations behind a $22$-declaration interface, compiling against Mathlib alone.}
\label{tab:lgen}
\end{table}

\medskip
The three phases reward \emph{different} strategies: a player who opens well can still
lose the endgame. The phase structure, not the win condition alone, is the substance of
the game.

\subsection{The three phases}\label{sec:phases}

\subsubsection{Early game: definitional scoping}
In the early game the player translates the source's target theorems into Lean
\emph{statements}, proofs left as \lean{sorry}. The difficulty is not the
statements but the \emph{definitions} they quantify over: every later move builds
on them, so a definition that bakes in the wrong hypothesis, metric, or regularity
class propagates its error through the whole development and is most expensive to
find late.

\begin{quote}\itshape
The early-game risk is definitional. The dominant strategy is to lock the
object scaffolding --- the definitions and the target signatures --- before
proving anything, and to read the source at the level of atoms (which exact
hypothesis, which exact moment, which exact norm), not of prose.
\end{quote}

The player commits the statements and their definitions, compiles with every proof
a \lean{sorry}, and only then proves. This ``API-lock'' separates committing to an
interface from discharging it --- and the interface is what every later phase is
written against, so errors caught here are cheap and the same errors caught in the
endgame are not.

\subsubsection{Mid game: steering the decomposition}
In the mid game the target's sorried statement is repeatedly \emph{decomposed} into
smaller sorried sub-statements, each heuristically easier, until the leaves are
within the agent's reach. The splitting itself is delegable --- a
\emph{sorry-decomposer} sub-agent does exactly that (Section~\ref{sec:casestudy})
--- which is what isolates the mid game's real demand: \textbf{steering was the
part of the loop that most demanded human judgment}. The agent
could prove a well-specified sub-lemma and split an oversized one; what it did not
supply was the judgment of \emph{where} to cut, which of several equivalent routes
the library could support, and when to abandon a route --- acts of mathematical
taste. The observation is time-stamped (this system, this moment; autonomous
steering is exactly the frontier systems improve at fastest), not a permanent
division of labor. In the case study the decisive steering was of this kind ---
which face of the distance to argue the stability estimate in, a call that meant
returning to the source's own formulation, and which form of its underlying
inequality the encoding could support (Section~\ref{ssec:dynamical-core}).

This is also where the loop \emph{learns}: a recurring pattern --- more often a
recurring \emph{failure} --- is distilled into a strategy and written into
$\mathcal{G}$, which is what makes the standing instruction file a live part of the
game rather than a fixed prompt.

\begin{quote}\itshape
The mid-game risk is choosing a decomposition the library cannot support, or one
that multiplies rather than reduces difficulty. The dominant strategy is
human-steered decomposition --- cut where the mathematician sees tractability ---
combined with encoding each hard-won heuristic into $\mathcal{G}$ so it is reused,
not re-derived.
\end{quote}

\subsubsection{End game: triaging the library gaps}
Eventually the surviving sorries are no longer decomposable into project-specific
pieces: they are requests for \emph{general} mathematics the ambient library may or
may not contain, and the endgame is the triage. Each falls into one of three kinds.
A \emph{phantom} looks like a gap but dissolves under a sharper reading of what the
consumer needs --- the case study had one, a conjectured metrization the
development never used (Section~\ref{ssec:dynamical-core}). A \emph{genuine gap you
build} is standard mathematics the library lacks that the player formalizes from
the primitives present --- here, a finite-dimensional duality theorem. A
\emph{genuine gap you defer} is marked with an explicit, cited placeholder when the
cost of building outweighs the value and the dependency is a recognized, true
theorem --- here, a Polish-space completeness fact for the Wasserstein metric,
deferred off every certified path.

\begin{quote}\itshape
The end-game risk is misclassifying: building infrastructure for a phantom, or
silently assuming a gap that should be marked. The dominant strategy is triage
against the library's actual coverage --- which requires knowing the library at
the functional-analysis level --- and verifying each candidate gap at the level
of what the consumer truly needs before committing to build it.
\end{quote}

Progress in formalizing research analysis is rate-limited by the ambient library's
infrastructure; the endgame is the work of locating, and either filling or honestly
marking, where it runs out.

\medskip\noindent
The case study of Section~\ref{sec:casestudy} left a complete, machine-readable record:
every move is a version-control commit, so the game's state --- its \lean{sorry} count, its
standing instruction file $\mathcal{G}$, its file structure --- can be reconstructed at any
point of its $362$-commit history.\footnote{Reproducible scripts and the raw data are in
the development's \lean{formalize/retrospective} directory.} Three conventions fix what
the figures count: ``live \lean{sorry}'' counts the token after stripping Lean line- and
block-comments (the word appears constantly in prose and docstrings, so an unstripped
count is meaningless; the count reaching $0$ at the final commit, where the build is
independently known clean, validates the strip); the production library is the Lean
package source, excluding the throwaway development scratch file; and the commit
classification of Section~\ref{ssec:timeframe} is a heuristic over commit-message intent.

\begin{figure}[t]
\centering
\includegraphics[width=\textwidth]{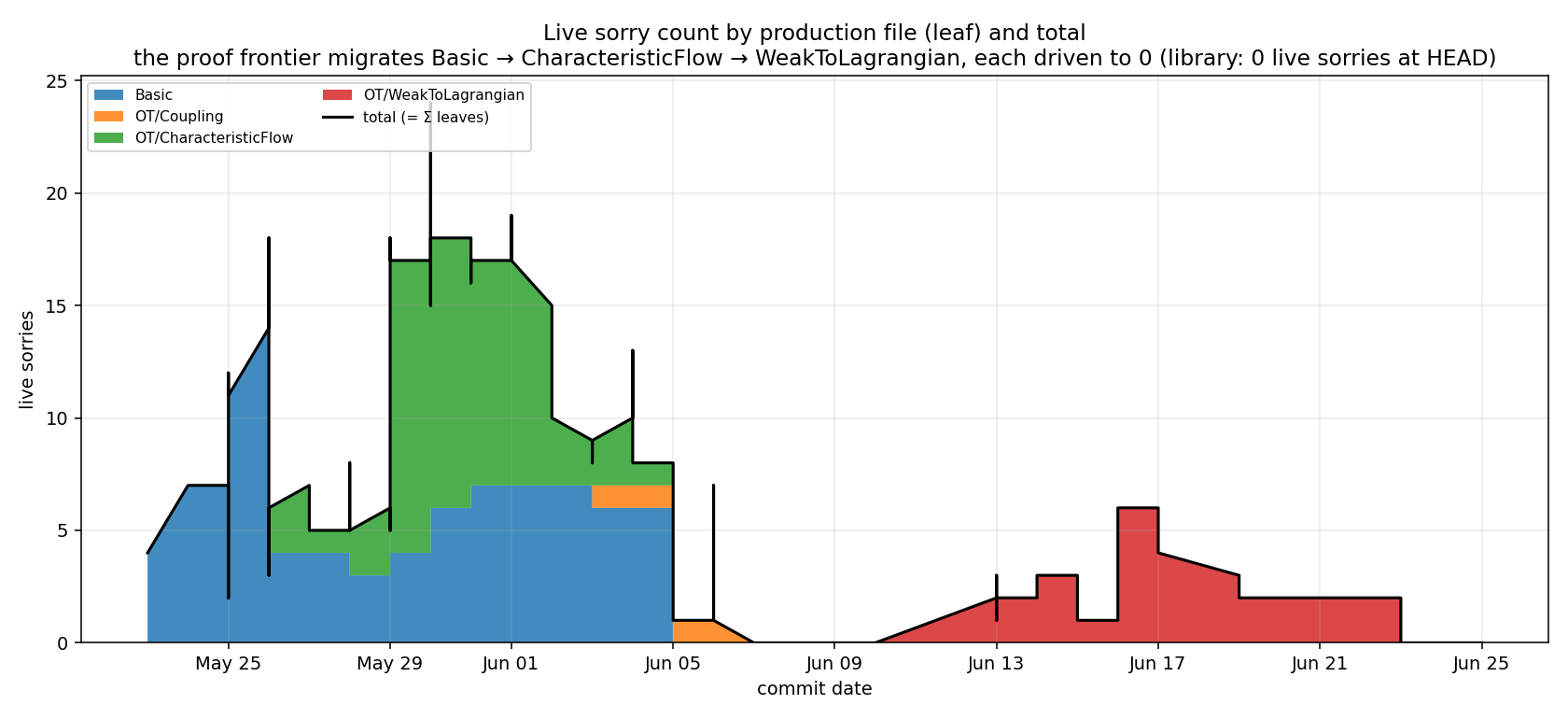}
\caption{From the case study of Section~\ref{sec:casestudy} ($362$ commits): live
\lean{sorry} count per production file, stacked so the leaves sum to the total (black
line). The open-goal front is concentrated in one active leaf at a time and migrates across
the development; the total peaks at $24$ and returns to $0$. The two humps are two games ---
the well-posedness/stability marquee and the superposition-principle follow-on ---
separated by the zero-\lean{sorry} reusability extraction.}
\label{fig:sorry}
\end{figure}

\needspace{10\baselineskip}
Three series, read off that record, turn the claims of this section and the next into data. The first is
the \lean{sorry} front: Figure~\ref{fig:sorry} plots the live count per source file, and
the three-phase rhythm is visible in it directly. Open goals stay concentrated in a single
\emph{active leaf} --- one production file at a time carries the bulk of the \lean{sorry}s,
and the front migrates from the basic objects, through the characteristic-flow
well-posedness core, to the weak-to-Lagrangian bridge, each interface locked and its goals
driven down before the front advances: the early-game API-lock discipline made visible. The
total never runs away, because the game is a sequence of bounded burndowns, not one
accumulating debt; and the flat stretch between the figure's two games, at zero live
\lean{sorry}, is the reusability extraction of Section~\ref{ssec:reuse}, banking the layer
without reopening a goal.

\subsection{The strategy framework}\label{sec:strategies}

The standing instruction file accumulated its lessons in four series, which operate
at genuinely different levels with different triggers for ``this rule applies.''

\begin{description}[leftmargin=2em,style=nextline]
\item[L-series --- Lean / agent-tool lessons.] Idioms, elaborator quirks,
parser quirks, and the engineering of agent sub-task specifications. Consulted
when writing or debugging an individual Lean proof.
\item[P-series --- Process discipline.] When to ask, commit, delegate, pivot, and
how to verify. Consulted at the per-move discipline level.
\item[M-series --- Mathematical structure.] Which mathematical structures to work
in, and what shape a proof should take. Consulted when designing a proof.
\item[B-series --- Bridging architecture.] When to build new infrastructure versus
patch an existing seam. Consulted when frictions accumulate across attempted bridges.
\end{description}
One representative lesson from each series --- earned, and reproduced verbatim from
$\mathcal{G}$ --- appears in Appendix~\ref{app:claudemd}.

The lessons were not designed in advance but \emph{earned} --- each keyed to a specific failure that
occurred during the game and was then encoded into $\mathcal{G}$ to prevent its
recurrence. The standing instruction file records the mathematician's accumulated strategic
experience in a form the agent can consult, amortizing the human's judgment across
the thousands of moves the agent makes.

\begin{figure}[!b]
\centering
\includegraphics[width=\textwidth]{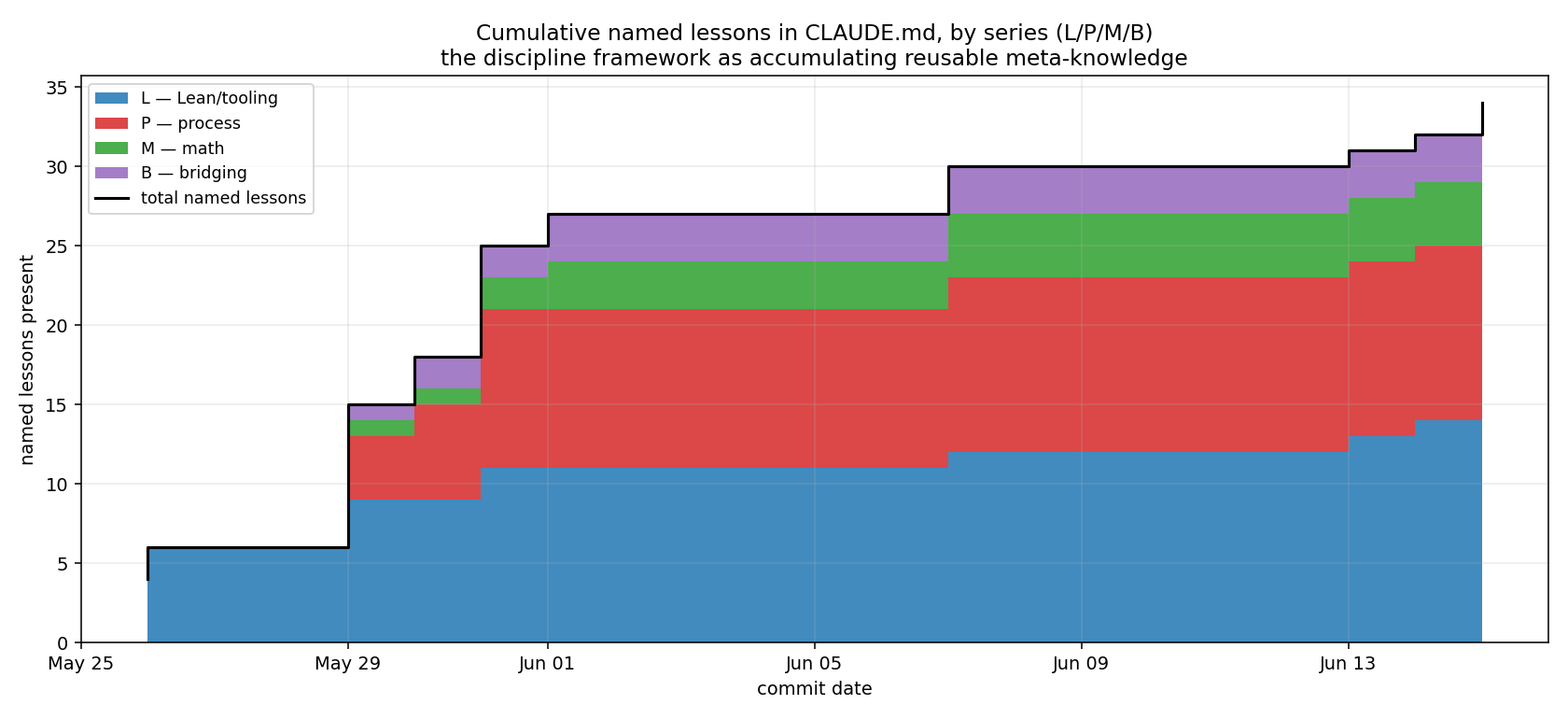}
\caption{From the case study of Section~\ref{sec:casestudy}: named strategies present in the
standing instruction file $\mathcal{G}$ across its $36$ revisions, by series (L/P/M/B). The
staircase shape --- flats punctuated by jumps at specific failures --- is the signature of
strategies earned rather than designed. The total grows $4\to34$; the four-series taxonomy
crystallizes in one revision, and the process (P) series jumps by four in the aftermath of
the cascade frictions of Section~\ref{sec:phases}.}
\label{fig:lessons}
\end{figure}

\medskip\noindent
The second series shows this accumulation as data: Figure~\ref{fig:lessons} plots the
count of named strategies in $\mathcal{G}$ across its $36$ revisions, split by series. Two
landmarks are legible: the four-series taxonomy crystallizes in a single revision (before
it, an undifferentiated list of tool lessons; after, all four series present), and the
process series jumps by four rules at once, in the aftermath of exactly the cascade of
frictions the three phases above describe. The closing growth is the tool-lesson series
alone: through the superposition follow-on it is Lean idioms that accumulate, not new
strategy, so the file tracks the novelty of the tactical terrain crossed rather than the
size of the proof obligation.

\section{The mean-field derivation, formalized}\label{sec:casestudy}

We played the game on Dobrushin's derivation \cite{Dobrushin} of the Vlasov
equation from $N$-particle Hamiltonian dynamics; for the broader mean-field theory
see \cite{Spohn,Golse,Neunzert}. This section replays the derivation in its
mathematical order --- get the objects right, build the distance, run the dynamical
core, reconcile the two solution notions, extract the layer; one subsection each ---
and at each stage marks the two things the finished proof hides: where the
mathematics was genuinely hard, and who unlocked it. The answer differs by stage ---
a human read, human-steered agent throughput, a human redirection, both at once --- and that
spread, visible in the derivation rather than asserted about it, is the division of
labor. The coupling argument of Section~\ref{ssec:dynamical-core} is shown in
Dobrushin's terms, at paragraph length --- it is where the optimal-transport
machinery is spent.
The endgame's library-building --- the general mathematics the ambient library
lacked --- is marked where it arises rather than gathered into a section of its own.

The agent system was Claude Code, Anthropic's agentic coding environment, driving a
frontier Claude model --- Opus 4.7, then 4.8, as the build spanned the upgrade. The
human and the model alternated moves in
an interactive session, and mechanical, well-scoped sub-tasks were delegated to
specialized sub-agents --- a \emph{sorry-prover} (close a single \lean{sorry} by
search against the local Mathlib), a \emph{sorry-decomposer} (split one oversized
\lean{sorry} into named helper lemmas), and library-scout and verifier agents ---
coordinated by a small driver script; these delegated sub-agents ran on
\lean{claude-sonnet-4-6} (recoverable from the project's agent logs). The standing
instruction file $\mathcal{G}$
of Section~\ref{sec:strategies} is the project's in-repository \lean{CLAUDE.md}. The
development is a public git repository, from which the exact model versions, the
full standing instructions, and the per-session agent logs are recoverable.

\subsection{From Newton to an exact weak solution}\label{ssec:newton-to-weak}

\noindent\emph{Hard: getting the solution concept right. Unlocked by: a human read of the test class.}

The derivation opens with $N$ identical particles in $\R^d$ under the mean-field
scaling, governed by the Newton equations
\[
  \dot{x}_i = v_i, \qquad
  \dot{v}_i = -\tfrac1N\textstyle\sum_{j\neq i}\nabla W(x_i-x_j),
  \qquad i=1,\dots,N,
\]
where the pair potential $W:\R^d\to\R$ is the section's standing assumption (the
type class \lean{AssW}): continuously differentiable, even, with globally Lipschitz
gradient of constant $L:=\mathrm{Lip}(\nabla W)$. The $1/N$ coupling holds the
kinetic and interaction energies at the same order as $N\to\infty$, and the bridge
to the kinetic picture is the \emph{empirical measure}
$\mu^N := \tfrac1N\sum_i\delta_{(x_i,v_i)}$, a probability measure on the phase space
$\R^d\times\R^d$ --- in Lean, \lean{empiricalMeasure} on \lean{PhaseSpace d}.

The mathematics of the opening is standard bookkeeping --- and, fittingly, among the
first sorries the agent closed. Differentiating $\langle\mu^N_t,\varphi\rangle$
along the Newton flow and \emph{adding and subtracting the diagonal} $i=j$ produces
the convolution against the spatial marginal plus a diagonal remainder, which the
evenness of $W$ (part of \lean{AssW}) kills: $\nabla W(0)=0$, so the empirical
measure solves the weak equation \emph{exactly}, not merely to $O(1/N)$. Even here
the win condition leaves a mark --- a remainder may not be discarded silently, so
the statement (\leanlink{weakEvolutionEmpiricalMeasure}) returns it explicitly, its
value and bound as conjuncts, and evenness collapses it in the corollary
\leanlink{empiricalMeasureSolvesVlasov}.

\needspace{12\baselineskip}
What was not routine is the definition the theorem quantifies over. The
weak-solution predicate ranges over the $C^\infty$ compactly supported test
functions:
\begin{leancode}
def IsVlasovSolution (gradW : PhysSpace d → PhysSpace d)
    (f : ℝ → Measure (PhaseSpace d)) : Prop :=
  ∀ φ : PhaseSpace d → ℝ, ContDiff ℝ (⊤ : ℕ∞) φ → HasCompactSupport φ →
    ∀ gradXφ gradVφ : PhaseSpace d → PhysSpace d,
      (∀ z, gradXφ z = gradient (fun x => φ (x, z.2)) z.1) →
      (∀ z, gradVφ z = gradient (fun v => φ (z.1, v)) z.2) →
      WeakEvolutionEq gradW f φ gradXφ gradVφ (fun _ => 0)
\end{leancode}
Its first version was \emph{definitionally wrong}, in a way that collapsed every
statement quantifying over it to a triviality. The error was ours, and it is the
early game's own point: getting a definition right at the start is hard, and the
wrong version typechecks. In the Mathlib we build against, the top exponent in
\lean{ContDiff ℝ ⊤} means \emph{real-analytic}, not $C^\infty$ (the meaning of
$\top$ had changed earlier in the library's history; the stale idiom still
compiles); a nonzero real-analytic function cannot have compact support, so the
test class had quietly collapsed to $\{0\}$ and the weak-solution hypothesis was
\emph{vacuously} true. The build stayed green throughout: the axiom footprint
certifies a proof relative to the statement as written, and cannot see that the statement had
become empty. A thirty-second \lean{\#check} caught the collapse, and the fix
(\lean{ContDiff ℝ (⊤ : ℕ∞)}, the $C^\infty$ element written above) restored the
class at twelve sites --- the clearest instance in the development of the boundary
Definition~\ref{def:win} draws.

\subsection{Wasserstein-1, in two faces}\label{ssec:wasserstein}

\noindent\emph{Hard: the bridge between the two faces. Unlocked by: human-steered search, agent throughput.}

The distance that organizes everything to follow is the Wasserstein-1 metric
$\Wone$ (Definition~\ref{def:w1}). The development carries it in two faces. The
\emph{dual} face is the definition: a supremum over Lipschitz test functions,
stated once over a general ground cost $c$ and specialized to $c=\mathrm{dist}$:
\begin{leancode}
def wassersteinCost (c : α → α → ℝ) (μ ν : Measure α) : ENNReal :=   -- DUAL
  ⨆ (f : α → ℝ) (_ : ∀ x y, |f x - f y| ≤ c x y),
    ENNReal.ofReal (∫ x, f x ∂μ - ∫ x, f x ∂ν)
def wasserstein1 (μ ν : Measure α) : ENNReal :=
  wassersteinCost (fun x y => dist x y) μ ν
\end{leancode}
The \emph{primal} face is an infimum over couplings of $(\mu,\nu)$
(Definition~\ref{def:w1}), and Kantorovich--Rubinstein duality equates the two
faces at $c=\mathrm{dist}$:
\begin{leancode}
def wasserstein1_coupling (μ ν : Measure α) : ENNReal :=            -- PRIMAL
  ⨅ (π : Measure (α × α)) (_ : IsCoupling π μ ν), ∫⁻ z, edist z.1 z.2 ∂π
theorem wasserstein1_eq_coupling … :                               -- KR duality
    wasserstein1 μ ν = wasserstein1_coupling μ ν
\end{leancode}

\medskip\noindent
The two faces are easy to write down; the \emph{bridge} between them had to be built,
and it is the reusable layer's most substantial construction. Its two pieces: the
coupling cost's triangle inequality
(\leanlink{wassersteinCost\_coupling\_triangle}), which disintegrates a coupling of
$(\rho,\nu)$ over $\rho$ and glues --- the load-bearing, error-prone part being not
the cost bound but that the glued measure has the \emph{correct marginals}, exactly
where a ``near-citation'' from the library hides a gap; and the hard direction of
the duality on finitely-supported measures
(\leanlink{wassersteinCost\_coupling\_le\_dual\_of\_finiteRange}), which the Mathlib
version we build against (v4.29.1) does not carry. The duality entered the
development as a marked, \lean{sorry}'d foundation and had to leave as the
endgame's built gap (Section~\ref{sec:phases}): the win condition certifies nothing
while the mark stands. The search for its proof was the mathematician's to steer.
The natural Hilbert-space form of the separation was assigned first, ran into
systemic friction with the inner-product instances on the Euclidean type, and was
called off; the appealing Birkhoff-vertex shortcut was scouted and called off (the
library's Birkhoff theorem is uniform-marginal; the transportation polytope here is
not); the third assignment survived --- the geometric separation on the plain
product $\iota\to\R$, assembling Mathlib's Farkas lemma (conic separation)
\cite{mathlib} with two compactness facts proved by hand (primal attainment on the
transport polytope; closedness of the image cone, the hypothesis Farkas consumes)
and a $c$-transform folding the dual pair into a single admissible potential --- an
\emph{inequality}, primal $\le$ dual, with no attained optimum anywhere.

Unlike the stages that turned on a single decision, this bridge was won by a
directed search plus throughput: the mathematician fixed the
target, assigned each avenue, and called each abandonment; the agent did the
sustained proof-work --- some twenty-five commits assembling Farkas, the compactness
facts, and the $c$-transform into the duality \leanlink{wasserstein1\_eq\_coupling},
the single optimal-transport line \lean{dobrushin} consumes. It is the clearest case
in the development of the machine carrying a hard, unglamorous build to the end.

\subsection{The dynamical core}\label{ssec:dynamical-core}

\noindent\emph{Hard: the stability estimate. Unlocked by: a human redirection --- back to Dobrushin's own coupling formulation.}

The limit equation is the nonlinear Vlasov equation, solved by the characteristic
flow $\dot X = V$, $\dot V = -(\nabla W*\rho_t)(X)$ transporting the datum $f_0$,
with $\rho_t$ the self-consistent spatial marginal --- the nonlinearity: the force is
generated by the very solution it transports. Its two marquee facts are
forward well-posedness at arbitrary Lipschitz constant $L$ (Theorem~\ref{thm:wp}),
a Banach fixed point in $\Wone$, and Dobrushin's estimate (Theorem~\ref{thm:dob}),
whose proof drives the mean-field limit. Both are won (Definition~\ref{def:win}).

A statement-level decision shaped the well-posedness theorem before any proof.
Dobrushin's is a \emph{forward} Cauchy problem, and the marquee is posed that way ---
existence on $[0,\infty)$ --- rather than as an $\exists!$ over all of $\R$. The
two-sided claim was too strong: it forced backward-time machinery the evolution does
not use. Uniqueness survives exactly
where the mathematics puts it --- per window, in the Lagrangian class
(\lean{vlasovWellPosedness\_uniqueness}, Theorem~\ref{thm:wp}) --- so nothing is lost
by proving forward. Matching the statement to what the argument supports is the early
game's other discipline, dual to definitional scoping.

Theorem~\ref{thm:wp} holds at \emph{arbitrary} $L$, and reaching that was a matter of the
\emph{proof}, not the statement. Dobrushin's global-in-time argument, tiled into short windows with a fixed unit
force-window, forces $L$ small ($L<1$) --- an artifact of the tiling, not of the
mathematics, since Dobrushin's theorem holds for every $L$. Tracing the dependency to its single structural origin (the window's unit width,
not any analytic mechanism --- the artifact-versus-genuine distinction of
Appendix~\ref{app:claudemd}) gave the fix: \emph{exact} tiling, windows of width exactly
$T/N$ so the chain lands on the target, killing the reach-slack a naive
offset-drop would leave. The removal was certified by instantiating the
theorem at $L=2$, a formerly forbidden value, and confirming it typechecks with no smallness
hypothesis.

\medskip\noindent
\textbf{The coupling argument.} The decisive move is a redirection. Read in the dual
$\Wone$, the estimate appears to need an \emph{optimal} coupling of the data, an
attainment theorem the library does not have. The mathematician's call was to return
to the source: Dobrushin states the estimate in the coupling cost from the start ---
his metric \emph{is} an infimum over couplings \cite{Dobrushin} --- and there it needs
only that the infimum lie below \emph{any} particular coupling's cost, no attainment.
The route is his; the judgment was the pairing of library gap with source formulation.

Fix \emph{any} coupling $\pi$ of the data, optimal or not. Each solution is
transported by its own characteristic flow, so pushing $\pi$ forward by the pair of
flows couples $(f_t,g_t)$, and \lean{wasserstein1\_pushforward\_le\_iInf} bounds
\[
  \Wone(f_t,g_t)\;\le\;\int \mathrm{dist}\!\big(\Phi^t_f z_1,\ \Phi^t_g z_2\big)\,
  d\pi(z_1,z_2)\;=:\;Q_t ,
\]
with $Q_0$ the cost of $\pi$ itself. Dobrushin's two estimates close the bound
\cite{Dobrushin}, transplanted from his truncated metric to the unbounded $\Wone$
with moment control: the force mismatch is controlled by the very quantity under
estimate (\lean{convolveFunctionMeasure\_lipschitz\_in\_x}), the trajectory gap obeys
the integral-form bound (\lean{flow\_difference\_mild\_bound}), and Grönwall
(\lean{gronwall\_mild\_le}) closes them to $Q_t\le Q_0\,e^{2Mt}$ with $M=\max(1,L)$
--- the rate $C=2\max(1,L)$ of Theorem~\ref{thm:dob}, the factor $2$ not cosmetic but
drift and forcing each contributing $M$. The mild form is Dobrushin's own, and it is
the form the encoding can support: Grönwall on $t\mapsto\Wone(f_t,g_t)$ directly ---
the development's first route --- demands a time-regularity the dual sup supplies
only as lower semicontinuity. The formalization runs on a finite window, each $t$
handled at $T=t+1$.

Taking the infimum over the initial coupling $\pi$ turns $Q_0$ into the primal
distance of the data, and one application of the Kantorovich--Rubinstein bridge
\lean{wasserstein1\_eq\_coupling} rewrites it as the dual $\Wone(f_0,g_0)$ of the
statement. That single \lean{rw} is the proof's \emph{only} appeal to duality: the whole
stability core, and the mean-field limit that follows from it, are built in the coupling
cost without it.

The mean-field limit follows: applying the estimate to
the Vlasov solution $f$ and the empirical curve gives, whenever the initial empirical
measures converge,
\[
  \sup_{t\in[0,T]} \Wone(\mu^N_t, f_t) \;\le\; e^{CT}\,\Wone(\mu^N_0,f_0)
  \;\xrightarrow[N\to\infty]{}\; 0,
\]
the separate corollary \lean{meanFieldLimit} (Corollary~\ref{cor:mfl}).

One conjectured foundation dissolved rather than being built. The development first carried a
metrization theorem relating narrow convergence and $\Wone$ as an external gap; its single
consumer needed only the easy lower-semicontinuity direction (Villani's, with moment control
\cite{Villani}), a short lemma from the dual representation. Recognizing the full metrization
as never load-bearing --- before building it --- is the endgame's triage in action: the
target theorems' footprint was already clean of it.

\subsection{The superposition principle, and why $C^2$}\label{ssec:superposition}

\noindent\emph{Hard: the superposition principle. Unlocked by: both --- a human typeclass decision, an agent construction.}

The construction of Section~\ref{ssec:dynamical-core} produces a solution of a
special form --- a pushforward of the datum along the characteristic flow --- and
Theorem~\ref{thm:wp}'s uniqueness lives in that Lagrangian class. Promoting it
toward \emph{all} weak solutions is the superposition principle
(Theorem~\ref{thm:superposition}), and what it turns on is, once
again, a \emph{definition} --- the regularity class of the potential. It costs
exactly one degree beyond \lean{AssW}; the two standing assumptions sit a single
field apart:
\needspace{9\baselineskip}
\begin{leancode}
class AssW (W : PhysSpace d → ℝ) : Prop where            -- C^{1,1}
  differentiable : Differentiable ℝ W
  even          : ∀ x, W (-x) = W x
  lipschitzGrad : ∃ L : NNReal, LipschitzWith L (fun x => fderiv ℝ W x)
class AssW2 (W : PhysSpace d → ℝ) extends AssW W : Prop where   -- C², one more field
  gradContDiff : ContDiff ℝ 1 (fun x => fderiv ℝ W x)   -- ∇W ∈ C¹, i.e. W ∈ C²
\end{leancode}

This is the one subsection where the Lean is not confirmation standing beside the
mathematics but the content itself. The proof reduces uniqueness for the nonlinear
equation to uniqueness for the \emph{linear} continuity equation with the field
frozen at $\rho_t$ --- both $f$ and the flow-pushforward solve it with the same
datum --- and proves the linear uniqueness by the \emph{dual transported test
function} $\psi(s,z):=\varphi(\Phi_{s\to T}(z))$, constant along characteristics.
For $\psi(s,\cdot)$ to be an admissible $C^1$ test function the flow must be $C^1$
in its initial point, and that single requirement is exactly what $\mathrm{AssW2}$
buys and $\mathrm{AssW}$ does not: at the bare $C^{1,1}$ the flow is only
bi-Lipschitz, the chain-rule identity holds merely Lebesgue-a.e.\ rather than
$f_t$-a.e., and the representation needs the full superposition theorem of Ambrosio
\cite{AGS}. (Dobrushin sidesteps the equivalence by taking the Lagrangian
representation as the \emph{definition} of solution, which is why the marquee
theorems need only \lean{AssW} and only this bridge pays for \lean{AssW2}.)

That $C^1$ dependence is the \emph{variational equation} --- absent from the
ambient library, supplied in house as
\leanlink{charFlow\_hasFDerivAt\_in\_initialPoint}: a $C^1$ input (the \lean{AssW2}
field), a \lean{HasFDerivAt} conclusion, the regularity ladder in one signature.
Formalization here did not merely \emph{certify} a known argument; it
\emph{sharpened} it, forcing the exact degree of smoothness out of the prose and
onto a type class, where a referee reads it off. The \emph{decision} that
superposition turns on exactly this, and hence on $\mathrm{AssW2}$, was the
mathematician's; the \emph{construction} --- a fundamental matrix
$\dot M = A(s,z)\,M$ by Picard iteration, then Fr\'echet differentiation under the
integral --- was some fifteen commits of mechanical proof the library did not
shorten, and the agent carried it: the division of labor at the resolution of a
single theorem.

\subsection{The reusability outcome}\label{ssec:reuse}

\noindent\emph{The outcome: the layer, extracted and certified.}

After the game was won, we reorganized the optimal-transport machinery into a
self-contained layer (Definition~\ref{def:selfcontained}), taking the layer to be
the \emph{reachable general core}: the general-mathematics declarations on a target
theorem's proof path. General infrastructure we built but did not use --- a deferred
truncated-metric regime and a few lemmas a refactor superseded --- is excluded: a
reusable layer should be what the development used, not everything general we
happened to build.

The cut, read off the declaration-level dependency graph, is one-directional: the
development outside the layer reaches it through a $22$-declaration interface ---
the Wasserstein-1 and transport-cost API, the coupling predicate \lean{IsCoupling},
and the two phase-space types --- and \emph{zero} edges run the other way.
Directionality and interface width, not the size of the layer, are what make it
upstreamable; the reverse-edge count is the hard fact a referee re-derives from
the graph.

Checks (Q2) and (Q3) turn that graph audit into artifacts a referee re-runs, and
both pass at \textsc{head}. Packaged as a standalone Lake project, the layer builds
against Mathlib alone --- no kinetic-theory file in its import graph (Q2) --- and
passes Mathlib's environment-linter suite ($16$ linters) with zero warnings (Q3).
The Q3 pass is annotated: seven declarations carry disclosed
\lean{@[nolint unusedArguments]} for interface arguments retained for caller
uniformity; eighteen genuinely-unused instance arguments were removed.

\needspace{12\baselineskip}
\noindent The whole certificate is one re-runnable block:
\begin{center}\small
\begin{tabular}{ll}
\hline
(Q1) reverse edges, general $\to$ specific & $0$ \\
(Q2) modules' standalone build vs.\ Mathlib alone & exit $0$ \\
(Q3) environment linters / warnings & $16$ / $0$ \\
(Q4) layer reachable from a target & $49/49$ \\
axiom footprint of the target theorems & \cleanfootprint{} \\
size: general / total declarations & $49/299 \approx 16\%$ \\
interface width $w$ & $22$ \\
\hline
\end{tabular}
\end{center}

\noindent The three (Q4) sub-checks differ in strength: reachability ($49/49$)
holds \emph{by construction} (the layer is the reachable core); generality is
\emph{certified by} (Q2) --- a layer module cannot compile against Mathlib alone if
it names a Vlasov object; non-redundancy is library-search \emph{evidence} (no
\lean{exact?} hit against the pinned Mathlib), not proof --- \lean{exact?} cannot
see a semantic duplicate stated in a different form.

The reported width is the honest one, not the smallest achievable: five superseded
declarations sit just outside the reachable core, and the interface names they still
use raise $w$ from a possible $17$ to $22$; pruning them is signature-touching ---
it re-triggers the full certification chain --- so $22$ stands. The size row,
likewise, is descriptive rather than part of the certificate (Definition~\ref{def:selfcontained}).

\section{Assessment and outlook}\label{sec:assessment}

\subsection{The verification discipline}\label{sec:discipline}

The discipline that keeps the game honest is a small set of standing rules, all
variations on a single theme: \emph{the machine-checkable artifact is the only
certification, and an automated agent's claim of success is not one.}

\begin{enumerate}[leftmargin=2em]
\item \emph{Read before building.} Every consequential
move is preceded by reading the actual construction at the atom level, not the
interface's stated summary. The phantom foundation and the tiling reach-slack
(Section~\ref{ssec:dynamical-core}) were both caught by reading the construction
rather than trusting a plausible summary of it.
\item \emph{Certify by the footprint, not the build.}
The win condition is the axiom footprint, and it is checked --- by hand,
re-running \printaxioms{} --- after every structural change, never inferred from
a green build or an agent's report.
\item \emph{A failed build poisons its checks.} A
secondary check that ran against cached artifacts from a failed build certifies
the old state; build success is confirmed before any check is trusted.
\item \emph{Move, verify, commit, one unit at a time.} Structural changes are
made one mathematical unit per step, each followed by build, footprint check, and
commit, so that any regression is localized to a single, revertible move. When a
move broke the build during the case study, the discipline caught it, reverted,
re-diagnosed, and redid the move --- and the procedure preventing that class of
error was encoded into $\mathcal{G}$.
\end{enumerate}

The discipline is also delegable in a specific way: the human can hand a
well-specified, mechanical sub-task to a background agent, provided the human
re-certifies the result by the footprint --- because the gates catch a failure,
but diagnosing and reverting it is a judgment the human retains. This
delegation-under-recertification is the operational form of the division of labor read off
the commit record next.

\subsection{Timeframe and the division of labor}\label{ssec:timeframe}

The well-posedness and stability marquee --- from the first translated theorem to the
certified stability estimate --- ran in about a week; the full development, with the
superposition follow-on, spanned about a month, on a flat-rate subscription ---
Anthropic's \$200-per-month Max plan, no per-token billing. The claim is not raw speed but
the \emph{mode} that makes it possible: a human mathematician as strategic director
over an AI execution engine, the division Section~\ref{sec:casestudy} marked stage
by stage.

\begin{figure}[t]
\centering
\includegraphics[width=0.86\textwidth]{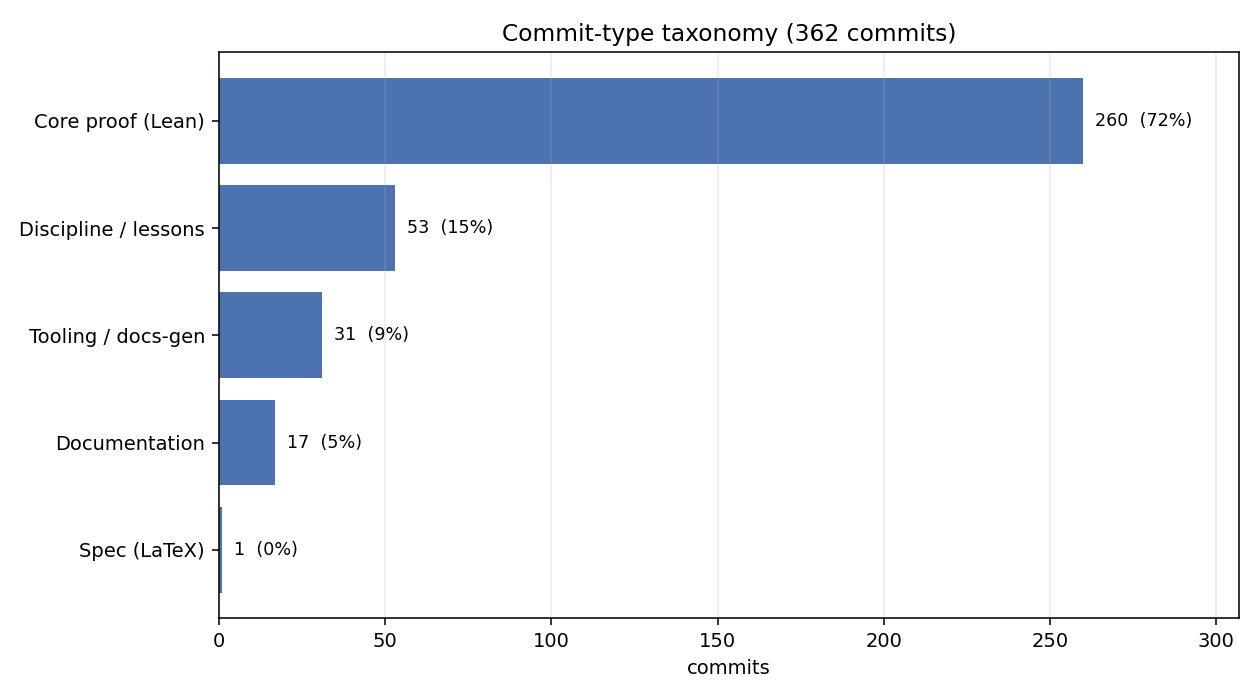}
\caption{From the case study of Section~\ref{sec:casestudy}: all $362$ commits classified by
primary intent. Roughly one in seven is pure meta-knowledge capture (discipline / lessons)
--- a direct, if heuristic, measure of the meta-learning loop, distinct from the proof work
itself.}
\label{fig:taxonomy}
\end{figure}

The commit record measures the
split. Figure~\ref{fig:taxonomy}, the third series, classifies all $362$ commits by primary
intent: just under three-quarters are core proof work, but roughly one commit in seven is
pure meta-knowledge capture --- the share of effort that goes not into proving the next
lemma but into distilling \emph{how} it should be proved (the meta-learning loop of
Section~\ref{sec:strategies}). The claim is that the division of labor, not either
party alone, is what the timeframe measures; the discipline above is why the speed
did not cost soundness.

We are not the only
data point: an independent, contemporaneous formalization \cite{IlinVML} --- a different
kinetic-theory target, the Vlasov--Maxwell--Landau equilibrium --- reports a comparable
mode, a single supervising mathematician completing the work in roughly ten days, at about
\$200, writing no Lean by hand. Two projects are not a controlled measurement, but they
make the strategic-director mode less of an outlier than a single run would
suggest.

\FloatBarrier

\subsection{What generalizes}

The three-phase structure and the strategy taxonomy (Section~\ref{sec:game}) are not
specific to kinetic theory: the definitional-scoping, decomposition-steering, and
library-gap-triage risks should recur in any formalization of research analysis,
where the rate-limiting resource is the ambient library's functional-analytic
coverage. Of the four strategy series, the P- and B-series are domain-independent,
the M-series carries analysis-flavored principles likely to recur in PDE
formalization, and the L-series is the most tool-specific and the likeliest to age
with the proof assistant. The transferable artifact is Appendix~\ref{app:claudemd}:
one verbatim lesson per series, liftable into another project's standing
instructions.

\subsection{Limitations}

The self-contained-layer check (Definition~\ref{def:selfcontained}) is pass/fail
rather than a number, and we have applied it to a single development;
the open question is not whether a fraction generalizes but whether its four
checks (Q1)--(Q4) transfer cleanly to other projects. Even in
this case the extracted layer still imports Mathlib wholesale rather than the
minimal files each declaration needs, so it remains a candidate for upstreaming
rather than a completed contribution. The
timeframe is a single observation that depends on the human director's prior fluency in both the mathematics and the
proof assistant. The strategy taxonomy was earned on one project and may be
incomplete; we expect later games to add series entries. Finally, one genuine
library gap in the case study (a Polish-space completeness fact for the
Wasserstein metric) was \emph{deferred} as a cited placeholder rather than built
--- it lies off the path of the certified target theorems; building it remains
future work, most naturally as an upstream contribution.

\subsection{Outlook}

The natural next edge is geometric. The Vlasov equation inherits a Hamiltonian,
Lie--Poisson structure from the $N$-particle system --- an energy $\mathcal{H}[f]$
and a bracket $\{F,G\}[f]$ for which $\partial_t f = \{f,\mathcal{H}\}$ --- and
identifying that bracket as the Lie--Poisson bracket on the dual of the Lie algebra
of Hamiltonian vector fields, arising as the $N\to\infty$ limit of the canonical
brackets on $(\R^d\times\R^d)^N$, is not formalized here; it is the subject of
\cite{MNPRS} and a natural target for a follow-on game. The present development
covers the dynamics but not the geometric structure of the limit. Marking that
boundary is the same statement-side discipline the paper runs on: say exactly what
is proved.

Beyond the next target is a practical question: how a working mathematician
should use these tools, given how fast they change. The premise is churn. The
agent system named in Section~\ref{sec:casestudy} will be superseded, perhaps
before this paper is refereed; the L-series lessons will age with the elaborator
they describe; the timeframe of Section~\ref{ssec:timeframe} is a single
observation about tools that no longer exist in that form. Nothing in this paper
should be read as advice about a particular system.

What ages more slowly is the craft. An analyst's working knowledge, beyond the
theorems, is a repertoire of moves --- integrate by parts and see what the
boundary term costs, split near field from far, trace a constant to see whether
it is structural or an artifact --- none of them theorems, all of them learned
from advisors and from failure, most of them older than the problems they are
currently applied to. The lessons of Section~\ref{sec:strategies} are knowledge
of the same kind, for a newer practice. They were earned the same way, and we
expect them to transfer the same way: not as a method to follow but as a
repertoire to draw on. We expect hybrid arrangements of this general shape ---
a mathematician directing, machines executing, the division shifting as the
machines improve --- to become a normal part of how analysis is done.

The game itself should age slowest of all, because its rules mention no
particular tool and no particular division of labor. The objective, the win
condition, and the layer are defined against the kernel and the ambient library;
\printaxioms{} does not care who made the moves. If the steering of
Section~\ref{sec:phases} migrates from the mathematician to the machine, the
game does not break --- the same checks score the new arrangement. That is what
the framing is for: a way to state what was attempted, what was won, and what
was left behind that stays meaningful while the tools underneath it turn over.

\paragraph{Acknowledgements.}
I thank the Stanford mathematics department for its hospitality and vibrant
community; Fred Rajasekaran for intriguing discussions about formalization;
Eleny Ionel for organizing last quarter's mathematics colloquium talks, which
kept us current with the latest developments in AI for mathematics; and the
Stanford Learning Seminar on Mathematics and Computation for its wonderful
seminars and discussions.

\appendix
\section{Formal statements}\label{app:formal}

The marquee results are stated as mathematics in the introduction
(Section~\ref{sec:statements}); their verbatim Lean signatures are collected here. An
ellipsis (\lean{…}) marks an elided standard binder or technical hypothesis, annotated
in an adjacent comment; the superposition principle's hypotheses are reproduced
complete. The
assumption classes \leanlink{AssW} and \leanlink{AssW2} and the two faces of the
Wasserstein-1 metric (\leanlink{wasserstein1}, \leanlink{wasserstein1\_coupling},
bridged by \leanlink{wasserstein1\_eq\_coupling}) appear in the body
(Sections~\ref{ssec:wasserstein} and~\ref{ssec:superposition}), where their exact
form is part of the argument.

\needspace{11\baselineskip}
\noindent The \emph{weak solution} predicate (Definition~\ref{def:weak-Vlasov}); its
test class is the subject of Section~\ref{ssec:newton-to-weak}:
\begin{leancode}
def IsVlasovSolution (gradW : PhysSpace d → PhysSpace d)
    (f : ℝ → Measure (PhaseSpace d)) : Prop :=
  ∀ φ : PhaseSpace d → ℝ, ContDiff ℝ (⊤ : ℕ∞) φ → HasCompactSupport φ →
    ∀ gradXφ gradVφ : PhaseSpace d → PhysSpace d,
      (∀ z, gradXφ z = gradient (fun x => φ (x, z.2)) z.1) →
      (∀ z, gradVφ z = gradient (fun v => φ (z.1, v)) z.2) →
      WeakEvolutionEq gradW f φ gradXφ gradVφ (fun _ => 0)
\end{leancode}

\needspace{11\baselineskip}
\noindent The \emph{Lagrangian solution} predicate (Definition~\ref{def:Lagrangian}):
\begin{leancode}
def IsLagrangianVlasovSolution (gradW : PhysSpace d → PhysSpace d)
    (f : ℝ → Measure (PhaseSpace d)) : Prop :=
  IsVlasovSolution gradW f ∧
  ∃ charX charV : ℝ → PhaseSpace d → PhysSpace d,
    IsCharacteristicFlow gradW (fun t => spatialMarginal (f t)) charX charV ∧
    (∀ t, f t = Measure.map (fun z => (charX t z, charV t z)) (f 0)) ∧
    (∀ s, AEMeasurable (fun z => (charX s z, charV s z)) (f 0))
\end{leancode}

\needspace{20\baselineskip}
\noindent \emph{Global well-posedness} and its per-window uniqueness (Theorem~\ref{thm:wp}):
\begin{leancode}
theorem vlasovWellPosedness (W : PhysSpace d → ℝ) [AssW W]
    (gradW : PhysSpace d → PhysSpace d) (hgradW : ∀ x, gradW x = gradient W x)
    (L : NNReal) (hL_gradW : LipschitzWith L gradW)
    (f₀ : Measure (PhaseSpace d)) (hf₀ : HasFiniteFirstMoment f₀) :
    ∃ f : ℝ → Measure (PhaseSpace d), f 0 = f₀ ∧
      (∀ t ∈ Set.Ici (0 : ℝ), HasFiniteFirstMoment (f t)) ∧
      (∀ T_target : ℝ, 0 < T_target →
        IsLagrangianVlasovSolutionOn gradW f T_target) ∧
      (∀ g : PhaseSpace d → ℝ, Continuous g → Bornology.IsBounded (Set.range g) →
        ContinuousOn (fun t => ∫ z, g z ∂f t) (Set.Ici 0))
theorem vlasovWellPosedness_uniqueness (W : PhysSpace d → ℝ) [AssW W]
    (gradW : PhysSpace d → PhysSpace d) (hgradW : ∀ x, gradW x = gradient W x)
    (L : NNReal) (hL : LipschitzWith L gradW) (hL_pos : (0:ℝ) < L)
    (f₀ : Measure (PhaseSpace d)) (hf₀ : HasFiniteFirstMoment f₀)
    {T : ℝ} (hT : 0 < T) (f g : ℝ → Measure (PhaseSpace d))
    (hf_init : f 0 = f₀) (hg_init : g 0 = f₀)
    …    -- finite moments on [0,T]; f, g both Lagrangian on [0,T]
    : ∀ t ∈ Set.Icc (0:ℝ) T, f t = g t
\end{leancode}

\needspace{14\baselineskip}
\noindent \emph{Dobrushin stability} (Theorem~\ref{thm:dob}):
\begin{leancode}
theorem dobrushin (W : PhysSpace d → ℝ) [AssW W]
    (gradW : PhysSpace d → PhysSpace d) (hgradW : ∀ x, gradW x = gradient W x)
    (L : NNReal) (hL : LipschitzWith L gradW)
    (f g : ℝ → Measure (PhaseSpace d))
    (hf : IsLagrangianVlasovSolution gradW f)
    (hg : IsLagrangianVlasovSolution gradW g)
    (hf_prob : ∀ t, HasFiniteFirstMoment (f t))
    (hg_prob : ∀ t, HasFiniteFirstMoment (g t)) :
    ∃ C : ℝ, 0 < C ∧ ∀ t, 0 ≤ t →
      wasserstein1 (f t) (g t) ≤
        ENNReal.ofReal (Real.exp (C * t)) * wasserstein1 (f 0) (g 0)
\end{leancode}

\needspace{17\baselineskip}
\noindent The \emph{mean-field limit} (Corollary~\ref{cor:mfl}):
\begin{leancode}
theorem meanFieldLimit (W : PhysSpace d → ℝ) [AssW W] (gradW …) (L) (hL …)
    (f₀ : Measure (PhaseSpace d)) (hf₀ : HasFiniteFirstMoment f₀)
    (f : ℝ → Measure (PhaseSpace d)) (hf_sol : IsLagrangianVlasovSolution gradW f)
    (hf_init : f 0 = f₀)
    (X V : (N : ℕ) → ℝ → Fin N → PhysSpace d)
    (hSol : ∀ N, IsNewtonSolution N gradW (X N) (V N))
    (hInit : Filter.Tendsto
      (fun N => wasserstein1 (empiricalMeasure N (X N 0) (V N 0)) f₀) atTop (nhds 0))
    (C : ℝ) (hC : 0 < C)
    (hDobrushin : ∀ N, DobrushinStabilityEstimate
      (empiricalMeasureCurve N (X N) (V N)) f C)
    (T : ℝ) (hT : 0 < T) :
    Filter.Tendsto (fun N => ⨆ t ∈ Set.Icc 0 T,
      wasserstein1 (empiricalMeasureCurve N (X N) (V N) t) (f t)) atTop (nhds 0)
\end{leancode}

\needspace{21\baselineskip}
\noindent The \emph{superposition principle} (Theorem~\ref{thm:superposition}), its
hypotheses reproduced complete --- nothing elided:
\begin{leancode}
theorem weak_isLagrangianVlasovSolutionOn
    (W : PhysSpace d → ℝ) [AssW2 W]
    (gradW : PhysSpace d → PhysSpace d) (hgradW : ∀ x, gradW x = gradient W x)
    (L : NNReal) (hL : LipschitzWith L gradW)
    (f : ℝ → Measure (PhaseSpace d)) (T : ℝ) (hT : 0 < T)
    (hf_weak : IsVlasovSolutionOn gradW f T)
    (hf_mom : ∀ t ∈ Set.Icc (0 : ℝ) T, HasFiniteFirstMoment (f t))
    (hf_narrow : ∀ (g : PhaseSpace d → ℝ), Continuous g → HasCompactSupport g →
      ContinuousOn (fun s => ∫ z, g z ∂(f s)) (Set.Icc 0 T))
    (hf_cont : ∀ x, Continuous
      (fun t => convolveFunctionMeasure gradW (spatialMarginal (f t)) x))
    (hf_cont_deriv : ContinuousOn
      (fun p : ℝ × PhysSpace d =>
        ∫ y, fderiv ℝ gradW (p.2 - y) ∂(spatialMarginal (f p.1)))
      (Set.Icc 0 T ×ˢ (Set.univ : Set (PhysSpace d))))
    (M_ρ : ℝ) (hM_ρ_nn : 0 ≤ M_ρ)
    (hM_ρ : ∀ t ∈ Set.Icc (0 : ℝ) T, ∫ y, ‖y‖ ∂(spatialMarginal (f t)) ≤ M_ρ)
    (hTL_PL : LocalSmallness_PL_buffer L T) :  -- L · T² < 1
    IsLagrangianVlasovSolutionOn gradW f T
\end{leancode}

\section{The standing instruction file: one lesson per series, verbatim}\label{app:claudemd}

The strategy framework is summarized in Section~\ref{sec:strategies} but
lives, in full, in the project's \lean{CLAUDE.md}. One entry from each of the four
series, lightly abridged, shows the form: each names a failure mode, its fix, and
the generalization that makes it reusable, and each was \emph{earned} from a
specific failure during the game.

\needspace{16\baselineskip}
\noindent An \emph{L-series} (Lean-idiom) lesson:
\begin{leancode}
### L10. `ring` requires `CommRing`; additive-group goals are `abel` territory

Failure mode: writing `by ring` on a purely additive goal — a difference such as
  -a - (-b) = b - a   or   (a + b) - (a + c) = b - c
— fails with `ring made no progress` when the underlying type is an `AddCommGroup`
but not a `CommRing` (e.g. EuclideanSpace ℝ (Fin d): pointwise + and -, but no
pointwise *).  The failure is opaque: the goal looks "obviously ring".

Fix: use `abel` (or `abel_nf`) for additive / vector-space / inner-product-space
goals; reserve `ring` for CommRing types (ℝ, ℂ, ℝ≥0, polynomial rings, …).

Operational rule: at the moment you write `by ring` on a vector-valued goal,
type-check "is this a CommRing?"  If not — and for phase-space differences it never
is — go straight to `abel`.  Avoids the build-fail-then-fix loop.
\end{leancode}

\needspace{14\baselineskip}
\noindent A \emph{P-series} (process) lesson:
\begin{leancode}
### P10. Build-permits vs. audit-certifies — a green build permits many stories

Failure mode: after a non-trivial refactor, a clean build is necessary but not
sufficient for "the refactor preserved meaning".  The build certifies typechecking
against the new types; it does not certify that consumers use those types for the
same mathematical purpose — a predicate doing double duty can split cleanly at the
signature layer while the body silently drops one use.

Operational rule: after a refactor that changes hypothesis semantics, do not
certify it "complete" on a green build alone.  The certifying instrument is
`#print axioms` — a body-level audit of the axiom footprint at each consumer.
Green build is a precondition, not a proof.
\end{leancode}

\needspace{18\baselineskip}
\noindent An \emph{M-series} (mathematical-structure) lesson:
\begin{leancode}
### M3/M4. Artifact smallness-constraints dissolve under the moving boundary

Failure mode: a smallness constraint (parameter < threshold) forced by a *static*
construction — a parameter made to bound its own growth — reads like genuine
mathematics but is an artifact of the construction, not of the theorem.  A green
build with the binder stripped does not certify its removal.

Diagnostic: trace the constraint to its unfold site.  If it makes a construction
parameter (a radius, a constant moment bound) large enough to contain its own
growth → artifact: dissolve it by rebuilding on the sharp time-local bound.  If it
makes a contraction ratio < 1 → genuine: carry it.

Fix: when removing a constraint imposed by chaining/tiling short pieces, use
*exact* tiling (windows of width T/N, landing on the target) — not a naive
offset-drop, which leaves a data-dependent reach-slack that looks general but fails
for large L.  Certify by instantiating the formerly forbidden value and confirming
it typechecks, not by a green build with the binder gone.
\end{leancode}

\needspace{12\baselineskip}
\noindent A \emph{B-series} (bridging-architecture) lesson:
\begin{leancode}
### B3. Conjunct shape: bounded below by the consumer, above by the producers

Failure mode: when strengthening a predicate with a new conjunct, reading only the
consumer's need fixes the *lower* bound and can leave the conjunct over-strong for
some producer to supply; reading only the producers can leave it under-strength.

Fix: read BOTH the consumer's genuine need and EVERY producer's available data
before fixing the conjunct — it belongs at the weakest-sufficient point in that
interval.  The one extra read (producer capacity) catches the
over-strong-but-unsupplyable shape before the edit, not mid-break.
\end{leancode}


\end{document}